# A Low-cost Humanoid Prototype Intended to assist people with disability using Raspberry Pi

Tariqul Islam Siam,

Mirzapur Cadet College

Email: tariqulislam.9@gmail.com

Md. Nayem Hasan Muntasir,

Mirzapur Cadet College

Email: n.moontasir@gmail.com

Md. Kamruzzaman Sarker,

University of Hartford

Email: sarker@hartford.edu

**Abstract**

This paper will try to delineate the making of a Humanoid prototype intended to assist people with disability (**PWD**). The assistance that this prototype will offer is rather rudimentary. However, our key focus is to make the prototype cost-friendly while pertaining to its humanoid-like functionalities. Considering growing needs of Robots, facilities for further installment of features have been made available in this project. The prototype will be of humanoid shape harnessing the power of Artificial Neural Network (ANN) to converse with the users. The prototype uses a raspberry pi and as the computational capability of a raspberry pi is minimal, we cut corners to squeeze the last drop of performance and make it as efficient as possible.

## 1 Introduction

The concept of humanoid Robots has thrilled humans ever since "Biomimetics" became a subject of study. [1] "Humanoid robots are professional service robots built to mimic human motion and interaction." [2] We delved deeper into the realm of integrating machines and flesh-bone humans rather than treating them as separate paradigms as the greater good lies in the marriage of nature and technology. Through the use of machines, we try to imitate human functionalities in support of their mode of operation. It might sound delusional to hear the term "low-cost humanoid". The word itself may be deemed as an oxymoron in a sense that humanoids tend to involve cutting-edge technology and quite a fascinating amount of engineering effort to make it feasible. Our team took the challenge to make a low-cost humanoid without ripping out most of its functionalities. **Vagabond** (the name we chose to call it) focuses on two key aspects; firstly, making it as cost-friendly as possible; secondly, enriching it with the features that might be of help to the people with disabilities. The main programming language used in this project is Python and for overall management, BASH scripts were also used to operate in a POSIX environment.

## 2 Equipment

Our Humanoid robot, herein referred to as Humanoid, needs very cheap equipment to build and to operate. The equipment we used:

| No. | Name | Price |
|---|---|---|
| 1 | Raspberry pi 4 model b 2GB [3] | 45$ |
| 2 | Mechanized Aluminum Body | 10$ |
| 3 | Jumper Wire [4] | 10$ |
| 4 | Servo Tower MG995 [5] | 7.73 x 8 = 61.84$ |
| 5 | Servo Tower Pro SG90 [6] | 1 x 4 = 4$ |
| 6 | Raspberry pi USB Microphone [7] | 14$ |
| 7 | Raspberry pi USB Speaker [8] | 14$ |
| 8 | Dot Matrix Display [9] | 4$ |
| | Total | 162.84$ |

**Table 1**: List of the components used (with market price)

## 3 Mode of Operation

Vagabond (Our Humanoid intended to assist PWD) prioritizes the ability to communicate with its users. A simple overview of its mode of operation is going to be elucidated here. A DNN [10] model is initiated at the very beginning and continues to run in the background regardless of any other activities being performed. It serves as the main coordinator of all processes and oversees all input-output operations performed by vagabond. Ensuring the presence of a trained chatbot (retrieval based) [11], the DNN then initializes the speech recognition function. Users' words are attempted to be recognized and made into string format using VOSK [12] offline speech recognition API. If a recognized command is found like "walk", "run", "pick up something", it invoke the task parser function realized by our team, the specific task is performed successfully provided that any termination command is not detected while performing that particular task. In case of no recognized command detection by the speech recognition function, it initiates the chatbot module, which is, in fact, a small DNN to converse with its users. The chatbot we made is of type "retrieval based". It preprocesses some predefined training data consisting of the parameters: tag, patterns, responses, context and feeds it to a 3 layers deep neural network. The neural network has been made generic to avoid overfitting complexities. We used 128 neurons in the first layer and only 64 neurons in the second layer. The third layer improvises its number of neurons based on the number of tags (genre of the piece of conversation) in the training data that means it must not exceed 32 (half of the previous layer) for optimum efficiency. It is to be noted that only three layers of neural network can continue basic conversation with accuracy up to 95 percent. We are also growing the training data and updating it frequently with conversations that happened with the bot earlier. So, we can deem the bot quite a bit intelligent and help the user to learn and become familiar with its mode of operation, recognized commands and their usage. The following recognized commands are thereby supplied by default with the humanoid.

Common Commands:
- Walk
- Run
- Stop
- Turn
- Pick up 'something'
- Home Assistant (requires internet access)
- Home Assistant mode supports multifarious instructions and is not limited to any particular command.

The home assistant feature requires internet access. We used PyWhatKit [13] module and extended its functions to materialize some basic home assistant features. So, the main code segments can be:
- Overseer
- Task Parser
- Chatbot
- Home Assistant

Now the code segments are going to be elucidated.

## 3 Overseer

The overseer function is mainly an umbrella for all the sub-functions. Basically, it itself is not that of import but rather its all-encompassing property is what we needed to manage all the code.

## 4 Task Parser for Body Movements

Task parser is a set of functions that mainly include the codes required to operate the mechanical body of the humanoidBasically, we spent 60 percent of our time making this humanoid to materialize the task parser functions such as walking, running, stopping, turning etc. Extensive study of the human walking mechanism was required to make the backbone of the walking code feasible. It was quite the ordeal. We used GPIO zero, pigpiofactory and the python threading module. GPIOzero was used to control the angular servo while pigpiofactory was used to convert the analog signals to PWM (Pulse Width Module) digital signals and thereby reduce servo jitters and improve walking stability. We coded a wrapper using the threading module and gained the capability to run multiple servos at a time while having independent control of them. Finally, the walking algorithm derived by observing human walking was implemented. The walking code is completely hard coded by our team and we are still improving it. The walking code in the task parser function serves as the backbone of all other functions. The task parser retains all the codes that are required to directly access the hardware. Overseer runs and manages them as the user requires it to be.

## 5 Chatbot (Retrieval Based)

If chatbot is in the limelight two of them get the floor i.e., Generative and Retrieval Based. Generative chatbots are resource heavy and provide less accuracy because of their property that they learn to generate unique responses based on template conversations provided to them. Retrieval based chatbot on the other had are light -weight and easy to design and implement on a minimal system. Our chatbot is trained on some basic conversation and response patterns to provide somewhat intelligent conversation capabilities without stressing the system resources too much. As retrieval based chatbots are quite simple a detailed description is not deemed necessary here. The properties of the chatbot are as follows:

| Name | Attribute |
|---|---|
| Type | Sequential |
| Activation | RELU, SoftMax |
| Optimization | SGD |
| Learning rate | 0.01 |
| Decay | 0.000001 |
| Loss | Categorical cross-entropy |
| Momentum | 0.9 |
| Dropout | 0.5 |
| Acceleration | Nesterov |
| Parameters | 24398 |
| Layers | 3 |
| Total Neurons | 128 + 64 + number of tags |

**Table 2**: Chart showing attributes and a summary of the properties of the neural network

## 6 Pre-processing

For pre-processing the training data, we used NumPy, NLTK (Natural Language Processing Toolkit) and JSON.

## 7 Home Assistant

If Vagabond is given access to the internet the user can use the home assistant feature. We used PyWhatKit and some other uncitable modules to code a simple home assistant. Our objective was to provide internet-based learning to PWD. Children with ASD can utilize this function and learn on their own using the internet without any kind of human intervention. In future, we hope to hone this feature as there are stations for further development in this field in our humanoid. As of now it covers the aspects like getting the date or asking for information on a particular topic language translation, brief summary of a particular topic, historical significance of current date etc.

## 8 How it helps the disabled

In our community, there is a school dedicated to educate children with special needs named **'Proyash[1]',** in Bengali, it means endeavor. After building the Humanoid at the lab, we went there and had a 5-day campaign where we tried to educate, enlighten, assist and most importantly, to have fun with these children with disabilities along with the humanoid. It is to be noted that our prototype was specially programmed to help them learn human communication without needing actual human involvement. Children with ASD generally have social-anxiety, severe introversion and shyness when speaking to humans [14]. If it's a robot, they can subdue their anxiety and are eager in general to know about the humanoid while implicitly developing their communication skills. The main perks of deploying a humanoid robot in assistance of children with ASD (Autism Spectrum Disorder) are:

1. Does not need active human involvement.

2. Produces good result in developing communication skills

3. The savants[15] will grow interest in humanoids and thereby can contribute to their future development.

4. As human involvement is not needed it produces great results in developing communication skills among children with ASD, social anxiety, trauma etc.

---

[1] The Name of a specialized school for children with Autism Spectrum Disorder

A graph showing communication performance improvement of children with ASD follows:

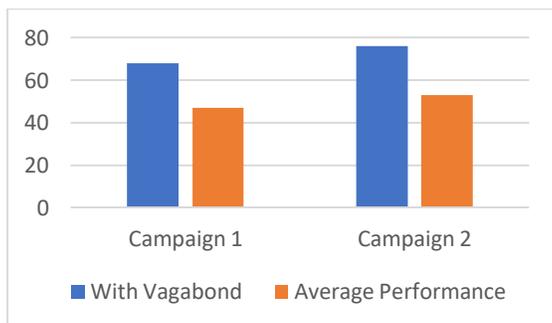

**Fig 1**: Chart showing improvement of communication score of children with **ASD** before and after the campaign. The scoring was done by "**Proyash**"[2] teachers themselves.

After a week with the vagabond, children were noticeably more responsive to stimuli and they were quite interested in that machine. Vagabond also looks a bit like toy; therefore, they also enjoyed it whilst implicitly developing their communication skills. The features they used the most are visualized in a chart herein:

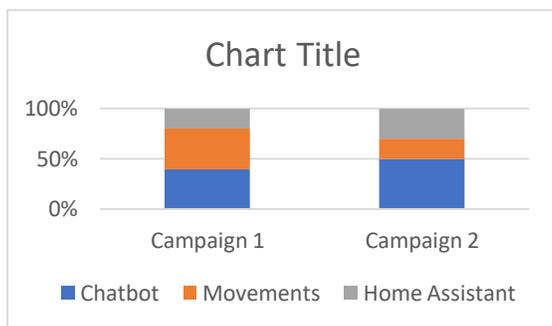

**Fig 2** : Chart depicting the inclination of children with ASD with respective features of the humanoid.

From the graph, it is clear that the children liked to use the chatbot feature very much. Generally, they do not get this type of non-human conversing agent so often, that's why, it has become their favorite. After our campaign, we got a fund equivalent of 300$. We made three more vagabonds and donated it to our local "Proyash" school with a view to improving the communication capabilities of the ASD children whilst remaining in their

---

[2] The Name of a specialized school for children with Autism Spectrum Disorder

comfort zone. It's a win-win for both. We hope to conduct more campaigns if funded properly.

## 9 Methodology

Vagabond derives its diverse functionalities from its flexible and resilient code structure. The code is divided into various segments to avoid any chance of critical failure. Interaction between different segments of code has been done through the operating system 'sys-calls'. It has the benefit of non-destructive editing in the code. Another benefit of this segmented code structure is that if one part becomes dysfunctional, the **overseer** can detect the error and subsequently report it to the developer while still running other functionalities regardless of the failure of the specific code. Discussing the code architecture and flow of operation is not paramount to this study but for clarification purposes, we will attempt to delineate it in the following.

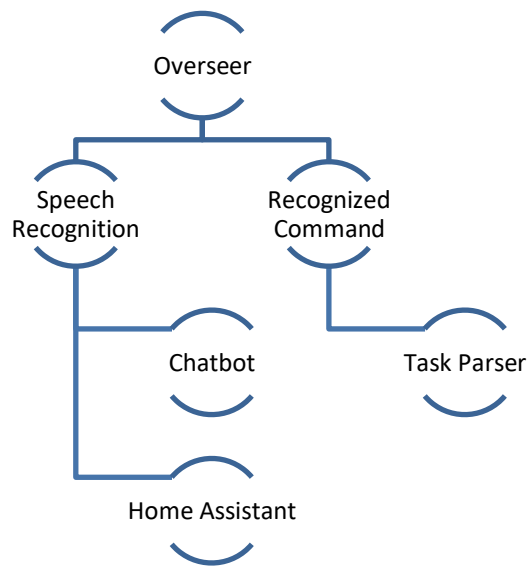

**Fig 3:** Flow chart depicting the code hierarchy of humanoid.

Speech recognition runs in a while loop continually in the background. It registers users' words and subsequently chats or helps as a home assistant. If a recognized command is found, then it passes the command to the task parser. The task parser then invokes a specific file for performing that particular operation. Input-output flow, piping etc. are done by the overseer code. Our code architecture is file based (codes are separated in files). This allows us to robustly manage all the diverse functionalities without making a mess and making it dysfunctional.

# 10 Conclusion

In this study our main objective was to develop something that can help persons with disability and simultaneously, be cost-effective. Vagabond was our primary step to achieving that. Our campaign showed great results in improving the communication skills of the ASD children. Without human intervention, they tend to perform better and being an interactive humanoid that can converse with children with quite a good accuracy, it gave them a great playful and frolicking company. It may be deemed as a cost-effective measure in assisting people with disability. Moreover, we can deploy this humanoid commercially to help the 'Disability Schools' which are remote, underdeveloped and lack human resources. In future, we expect to further develop the functionalities of vagabond and thereby hope our study may contribute to the society and be helpful to the children with special needs.